\documentclass[letterpaper, 10 pt, conference]{ieeeconf}  

\IEEEoverridecommandlockouts                            
\usepackage{graphicx,gen_settings}
\usepackage{tikz}
\usetikzlibrary{calc,arrows,arrows.meta,automata,positioning,backgrounds}
\usetikzlibrary{fit}
\usepackage{makecell}
\usepackage{epstopdf}
\usepackage{pgfplots}
\usepackage{pgfplotstable}
\usepackage{siunitx}
\usepackage{amsmath}
\usepackage{amssymb, url}
\usepackage[caption=false]{subfig}
\usepackage{cite}

\definecolor{TUMblue}{rgb}{0.00, 0.40, 0.74}
\definecolor{TUMgray}{rgb}{0.85, 0.85, 0.86}
\definecolor{TUMpantone285C}{rgb}{0.00, 0.45, 0.81}
\definecolor{lightblue}{rgb}{0.7529,0.8118,0.9333}
\definecolor{TUMpantone158}{RGB}{227, 114, 34}
\definecolor{TUMpantone383}{RGB}{162, 173, 0}
\definecolor{TUMgray}{RGB}{51,  51,  51}
\usepackage[nohyperlinks, nolist]{acronym}
\usepackage{microtype}
\usepackage{mathtools,xparse}
\usepackage{algorithmic,algorithm}
\DeclarePairedDelimiter{\norm}{\lVert}{\rVert}
\newtheorem{definition}{Definition}
\newtheorem{example}{Example}
\newtheorem{corollary}{Corollary}

\title{\LARGE \bf
Safe Reinforcement Learning with Probabilistic Guarantees \\ Satisfying Temporal Logic Specifications in Continuous Action Spaces
}

\author{Hanna Krasowski, Prithvi Akella, Aaron D. Ames, and Matthias Althoff
\thanks{The authors gratefully acknowledge the partial financial support of this work by the research training group ConVeY funded by the German Research Foundation under grant GRK 2428, by the project TRAITS funded by the German Federal Ministry of Education and Research, and by an IFI scholarship funded by the DAAD. Prithvi Akella was supported the Air Force Office of Scientific Research, grant FA9550-19-1-0302, and the National Science Foundation, grant 1932091.}
\thanks{H. Krasowski and M. Althoff are with the Technical University of Munich, Munich, Germany
        {\tt\small \{hanna.krasowski, althoff\}@tum.de}}%
\thanks{H. Krasowski, P. Akella and A. D. Ames are with the California Institute of Technology, Pasadena, USA
        {\tt\small \{pakella, ames\}@caltech.edu}}%
\thanks{© 2023 IEEE.  Personal use of this material is permitted.  Permission from IEEE must be obtained for all other uses, in any current or future media, including reprinting/republishing this material for advertising or promotional purposes, creating new collective works, for resale or redistribution to servers or lists, or reuse of any copyrighted component of this work in other works.} %
}

\begin{document}

\begin{acronym}
\acro{rl}[RL]{Reinforcement Learning}
\acro{mdp}[MDP]{Markov decision process}
\acro{stl}[STL]{Signal Temporal Logic}
\acro{ltl}[LTL]{Linear Temporal Logic}
\end{acronym}
\newcommand{\todo}[1]{\textcolor{red}{!ToDo! \quad #1}}
\newcommand{\vSRL}{safe \ac{rl} }
\newcommand{\VSRL}{Safe \ac{rl} }
\newcommand{\vSRl}{safe \ac{rl}}

\maketitle
\thispagestyle{empty}
\pagestyle{empty}

\maketitle

\begin{abstract}%
 Vanilla Reinforcement Learning (RL) can efficiently solve complex tasks but does not provide any guarantees on system behavior. To bridge this gap, we propose a three-step safe RL procedure for continuous action spaces that provides probabilistic guarantees with respect to temporal logic specifications.  First, our approach probabilistically verifies a candidate controller with respect to a temporal logic specification while randomizing the control inputs to the system within a bounded set. Second, we improve the performance of this probabilistically verified controller by adding an RL agent that optimizes the verified controller for performance in the same bounded set around the control input. Third, we verify probabilistic safety guarantees with respect to temporal logic specifications for the learned agent. Our approach is efficiently implementable for continuous action and state spaces. The separation of safety verification and performance improvement into two distinct steps realizes both explicit probabilistic safety guarantees and a straightforward RL setup that focuses on performance. We evaluate our approach on an evasion task where a robot has to reach a goal while evading a dynamic obstacle with a specific maneuver. Our results show that our safe RL approach leads to efficient learning while maintaining its probabilistic safety specification.
\end{abstract}

\section{Introduction}
\label{sec:introduction}
\ac{rl} has the potential to solve intricate tasks by learning complex policies efficiently. However, vanilla \ac{rl} cannot provide (probabilistic) safety guarantees, which is essential for real-world applications. Formal methods can eliminate this problem when integrated into the learning process. 
The most prominent formal methods approaches achieving safety guarantees for \ac{rl} with continuous action spaces are control-theoretic methods such as model predictive control~\cite{Gros2020.safetyshield, Wabersich2021a.safetyshield}, control barrier functions~\cite{Cheng2019.safetyshield}, or reachability analysis~\cite{Akametalu2014safetyshield, Kochdumper2022.safetyshield, Shao2021.safetyshield}. However, all these methods only handle reach-avoid safety specifications since they either determine unsafe state sets and prohibit the \ac{rl} agent from entering them or determine safe state sets and force the \ac{rl} agent to remain within those. 
Other methods are needed whenever the safety specification is more complex and cannot be seamlessly translated into a reach-avoid problem.

One way of expressing more complex safety specifications is via temporal logic. Indeed, there has been significant work that combines \ac{rl} with logical specifications for discrete action spaces \cite{Alshiekh2018.safetyshield, Konighofer2020safetyshield, Hasanbeig20.TLwithRL}. For example, Alshiekh et al.~\cite{Alshiekh2018.safetyshield} filter all unsafe actions proposed by the \ac{rl} agent with a safety shield synthesized from linear temporal logic specifications. 
Hasanbeig et al.~\cite{Hasanbeig20.TLwithRL} include the temporal logic specifications in the \ac{rl} process through a pessimistic and an optimistic learner where the pessimistic learner limits the exploration to low-risk actions.  
Still, applying approaches for discrete action spaces to real-world systems with continuous control inputs requires a low-level controller that converts the discrete actions to continuous inputs.
\begin{figure*}[t]
	\vspace{0.2cm}
	\centering
	\resizebox{0.99\textwidth}{!}{%
		\input{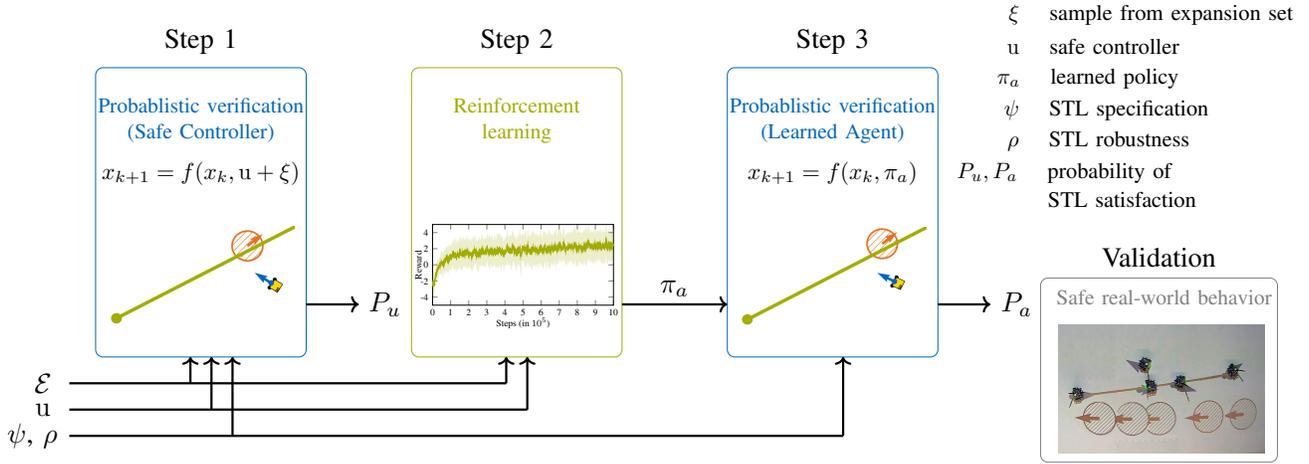}
	}
    \caption{\VSRL process for STL safety specification $\psi$ with robustness measure $\rho$ using a safe controller $\mathrm{u}$ and a system model ${x}_{k+1} = f(x_k, \cdot)$ where $\cdot$ can be replaced by any controller.}
    \label{fig:overall_pipeline}
	\vspace{-0.2 in}
\end{figure*}

Other \ac{rl} approaches realize continuous actions and guide the agent by a temporal logic specification that includes safety and performance objectives~\cite{Lavaei20.TLControlSynthesis, Gundana2021.TLwithRL, Cai2021.TLwithRL}, \textit{i.e.}, the temporal logic specification describes the entire task. For example, Cai et al.~\cite{Cai2021.TLwithRL} transform linear temporal logic into a B\"uchi automaton, which is integrated into \ac{rl} and yields probabilistic guarantees. 
Although specifying the task via temporal logic for \ac{rl} is promising, safety and performance objectives included in the task are potentially not aligned.
Possible solutions are to provide feedback to the user whenever the temporal logic specification becomes infeasible~\cite{Gundana2021.TLwithRL} or to trade off between safety and performance~\cite{Cai2021.TLwithRL}. 
The first solution is usually not practical for autonomous real-world systems, and the second one does not provide an explicit probabilistic guarantee for the safety specification, which might be required. 
Instead, our approach separates safety and performance objectives such that we obtain probabilistic safety guarantees while the feasibility is ensured by utilizing a probabilistically verified safe controller.  

To provide probabilistic safety guarantees, we leverage existing work in the probabilistic verification literature taking a scenario approach to risk-aware probabilistic verification~\cite{akella2022sample,weng2022safety}.  Here, the standard approach as described in~\cite{corso2021survey} is to pose verification as an optimization problem minimizing a quantifiable satisfaction measure provided by either a temporal logic specification or another method.  

\newidea{Contribution:} We propose a three-step safe \ac{rl} approach that improves the performance of a probabilistically verified black-box controller and results in probabilistic safety guarantees for the learned agent. 
Our key idea is to separate safety and performance objectives in distinct steps (see Fig.~\ref{fig:overall_pipeline} with probabilistic verification steps for safety and \ac{rl} for performance), which leads to efficient \ac{rl} while providing explicit safety guarantees.
In contrast to existing methods, our approach is suited for complex real-world systems since it is tailored to continuous action spaces, can probabilistically verify arbitrary \ac{stl} specifications, and does not require a system model to predict the safety of future actions. 
We validate our approach on an evasion task in a simulated dynamic environment and show that it can be translated to real-world systems as demonstrated through experiential results on the Robotarium~\cite{Pickem2017}.

\newidea{Structure:} The remainder of this paper is organized as follows. First, we introduce preliminary concepts in Sec.~\ref{sec:preliminaries}. Second, we present our \vSRL approach in Sec.~\ref{sec:approach}. Then, we explain the details of our safe evasion task and its experimental validation in Sec.~\ref{sec:casestudy}. Finally, we discuss our approach in Sec.~\ref{sec:discussion} and conclude in Sec.~\ref{sec:conclusion}. 

\section{Preliminaries}\label{sec:preliminaries}
\newidea{Signal Temporal Logic} \ac{stl} is a language by which rich, time-varying system behavior can be expressed succinctly and concisely. \ac{stl} is based on predicates $\tau$ which are Boolean-valued functions taking a truth value for each state $x \in \mathcal{X}$.  Predicates $\tau$ and specifications $\psi$ are defined in Backus-Naur notation~\cite[Section~2.1]{Bartocci2018} with respect to predicate functions $h_{\tau}$ that define subsets of a state space $\mathcal{X}$ where $\tau$ evaluates to $\true$:
\begin{gather}
    \label{eq:predicate_def}
    \tau(x) = \true \iff h_{\tau}(x)\geq 0,~h_{\tau}: \mathcal{X} \to \mathbb{R}, \\
    \label{eq:spec}
    \psi \triangleq
    \tau~|~\neg \psi~|~\psi_1 \lor \psi_2~|~
    \psi_1 \until_{[a,b]} \psi_2,
\end{gather}
where, $\psi \in \mathbb{S}$, and $a,b\in\mathbb{R}_{\geq 0} \cup \{\infty\},~b\geq a$. Here, $\mathbb{S}$ is the set of all \ac{stl} specifications which are evaluated over signals $s: \mathbb{R}_{\geq 0} \to \mathbb{R}^n$, and the space of all signals $\signalspace = \{s~|~s:\mathbb{R}_{\geq 0} \to \mathbb{R}^n\}$.  Finally, we denote that a signal $s$ satisfies $\psi$ at time $t$ via $(s,t) \models \psi$.  
Furthermore, every \ac{stl} specification $\psi$ has a \textit{robustness measure} $\rho$~\cite{maler2004monitoring}:
\begin{definition}
\label{def:robustness}
A function $\rho: \signalspace\times~\mathbb{R}_+ \to \mathbb{R}$ is a \textit{robustness measure} for an \ac{stl} specification $\psi$ if it satisfies: $\rho(s,t) \geq 0 \iff (s,t) \models \psi$.
\end{definition}
\begin{example}
Let $\psi = \neg(\true \until_{[0,2]} |s(t)| > 2)$, then any real-valued signal $s:\mathbb{R}_{\geq 0} \to \mathbb{R}$ satisfies $\psi$ at time $t$, \textit{i.e.} $(s,t) \models \psi$ if $\forall~t' \in [t,t+2],~|s(t')| \leq 2$.  A possible robustness measure is $\rho(s,t) = \min_{t' \in [t,t+2]}~2-|s(t')|$.
\end{example}
\noindent Note that while defining a robustness measure as per Definition~\ref{def:robustness} aligns with prior works~\cite{lindemann2018control,lindemann2020barrier} and our predicate definition in \eqref{eq:predicate_def}, it is not the only way of defining such a measure, \textit{e.g.} see Definition 3 in~\cite{donze2010robust} or Section 2.3 in~\cite{fainekos2009robustness}. 

\spacing
\newidea{Probabilistic Controller Verification}
As expressed in Section 3 in~\cite{corso2021survey}, \ac{stl} provides a natural way of phrasing black-box controller verification as an optimization problem over a space of parameters $p \in \mathcal{P}$ affecting signal generation.  More specifically, let $\mathcal{P}$ be a space of parameters denoting different environmental states in which we expect our closed-loop system to operate. For example, for warehouse robotics, these environmental states could be package and drop-off locations, the floor plan, \textit{etc}.  Additionally, for any specific environment parameter $p$, there may exist disturbances affecting system behavior.  Thus, we expect the closed-loop trajectory $\phi_p$, which is realized by our system for a specific environment parameter $p$, to be a sample of a $p$-parameterized random variable $\Phi_p$ with corresponding distribution $\pi_p$. 
\noindent To formulate the theorem used in this work, let us introduce $\uniform[\cdot]$ as uniform distribution and $\prob_\triangle$ denoting a probability where $\triangle$ indicates the underlying distribution.
\begin{theorem}
    \label{thm:prob_verification}
    (Adapted from~\cite[Thm. 7]{akella2022sample}) Let $\mathcal{P}$ be a space that admits a uniform distribution and let $\mathcal{D} = \{r_i = \rho(\phi_{p_i})\}_{i=1}^N$ be a set of $N$ closed-loop system robustnesses $r_i$, evaluating the robustness of one closed-loop trajectory sample $\phi_{p_i}$ per i.i.d. sample $p_i$ drawn from the uniform distribution over $\mathcal{P}$.  Furthermore, define $\rho^*_N = \min\{r_i \in \mathcal{D}\}$.  For any $\epsilon \in [0,1]$, the probability that $\rho^*_N$ underperforms the $1-\epsilon$-th quartile of possible robustness values is bounded below by $1-(1-\epsilon)^N$, \textit{i.e.} with $\mu \triangleq \uniform[\mathcal{P}] \times \pi_p$,
    \begin{equation}
        \label{eq:prob_statement}
        \prob^N_{\mu}\left[\prob_{\mu}[\rho(\phi_p) \geq \rho^*_N] \geq 1-\epsilon \right] \geq 1-(1-\epsilon)^N.
    \end{equation}
\end{theorem}

For a more detailed derivation of this theorem and its implications on dimensional scaling, we refer the interested reader to~\cite{akella2022sample}.

\newidea{Overarching Problem Statement:}
We assume that we have a safety specification $\psi$ expressed in \ac{stl} and an associated robustness measure $\rho$ as per Definition~\ref{def:robustness}. We also assume that a model and black-box controller $\mathrm{u}$ are available:
\begin{equation}
\begin{split}
    \label{eq:basic_information}
    x_{k+1}& = f(x_k,u_k),~x_{k},x_{k+1} \in \mathcal{X} \subseteq \mathbb{R}^n,\\ & u_k \in \mathcal{U} \subseteq \mathbb{R}^m,~ \mathrm{u}: \mathcal{X} \to \mathcal{U},
\end{split}
\end{equation}
where $\mathrm{u}(x_k) = u_k$ and the subscripts denote the time step. Additionally, we assume that the distribution of system behavior $\pi_p$ is time-invariant. Note that we do not need to explicitly know the model $f$ but can also employ a black-box simulation environment. 
The problem is to ensure probabilistic safety guarantees specified via \ac{stl} for the given system while using \ac{rl} for improving the performance.

\section{Safe RL Process}\label{sec:approach}
\newidea{Step One:} Our safe RL concept consists of three steps as shown in Fig.~\ref{fig:overall_pipeline}. Our first step is to follow the probabilistic verification procedure outlined in Sec.~\ref{sec:preliminaries} to verify whether the controller $\mathrm{u}$ realizes safe behavior when adding the uniformly sampled disturbance $\xi$: 
\begin{equation}
    \label{eq:stochastic_model}
    x_{k+1} = f(x_k,\mathrm{u}(x_k) + \xi),\quad \xi \sim \uniform[\mathcal{E}],~ \mathcal{E} \subseteq \mathbb{R}^m.
\end{equation}
Here, $\uniform[\mathcal{E}]$ corresponds to the uniform distribution over the set $\mathcal{E}$, which we assume to be fixed and independent of the system state $x \in \mathcal{X}$.  Per Theorem~\ref{thm:prob_verification}, 
we can determine the following probabilistic lower bound on the robustness measure value achievable by the closed-loop system~\eqref{eq:stochastic_model}.
\begin{corollary}
    \label{corr:prob_verify_cand_controller}
    Let the system dynamics be as per~\eqref{eq:stochastic_model}, the safety specification $\psi$ have robustness measure $\rho$ as per Definition~\ref{def:robustness}, and $\mathcal{D} = \{r_i = \rho(\phi_{x^i_0})\}_{i=1}^N$ be the robustnesses of $N$ trajectories $\phi_{x^i_0}$ where the initial conditions $x^i_0$ were uniformly sampled over $\mathcal{X}$.  Define $\rho^*_N = \min\{r_i \in \mathcal{D}\}$, then for some $\epsilon \in [0,1]$, $\rho^*_N$ underperforms the $1-\epsilon$-th quartile robustnesses achievable by the stochastic closed-loop system in~\eqref{eq:stochastic_model} with minimum confidence $1-(1-\epsilon)^N$, \textit{i.e.} with $\mu = \uniform[\mathcal{X}]\times \uniform[\mathcal{E}] \times \uniform[\mathcal{E}] \times \dots$,
    \begin{equation}
        \prob^N_{\mu} \big[\prob_{\mu} \left[\rho(\phi_{x_0}) \geq \rho^*_N \right] \geq 1-\epsilon \big] \geq
        1-(1-\epsilon)^N.
    \end{equation}
\end{corollary}
\begin{proof}
    This is an application of Theorem~\ref{thm:prob_verification}.
\end{proof}

\noindent Following Corollary~\ref{corr:prob_verify_cand_controller}, we can identify the robustness set $\mathcal{D}$ by sampling $N$ trajectories of the closed-loop system with controller $\mathrm{u}$ under bounded input disturbance from $\mathcal{E}$. Given the robustness set $\mathcal{D}$ and a specified $\epsilon$, we can evaluate the probabilities in Corollary~\ref{corr:prob_verify_cand_controller} and obtain $\rho^*_N$. We abbreviate this probabilistic verification process with the function $\mathrm{probv}(f, \mathrm{u}, \mathcal{E}, N, \epsilon)$. The first step concludes once we verified that a candidate controller $\mathrm{u}$ achieves safe behavior with a high probability, \textit{i.e.} $\rho^*_N \geq 0$ with $1-\epsilon \approx 1$. 

\newidea{Remark on Maximizing $\mathcal{E}$:} The second step in our \vSRL process will be to learn a policy that chooses disturbances within the expansion set $\mathcal{E}$.  As such, maximizing the size of this set has a direct impact on the ability of our procedure to learn a performant policy.  To that end, we propose a possible algorithm to expand $\mathcal{E}$ in Alg.~\ref{alg:disturbanceset}.

\begin{algorithm}
 \caption{findExpansionSet()}
 \label{alg:disturbanceset}
 \begin{algorithmic}[1]
 \renewcommand{\algorithmicrequire}{\textbf{Input:}}
 \renewcommand{\algorithmicensure}{\textbf{Output:}}
 \REQUIRE executable system $f$, controller $\mathrm{u}$, initial expansion interval set $\mathcal{E}_{init}$, vector of fractions to increase set \\ $\Delta \mathbf{f} \in \mathbb{R}^m$, number of samples $N$, quartile parameter $\epsilon$ \\
 \ENSURE Final expansion set $\mathcal{E}$
 \STATE $i = 1$
 \STATE $\rho^*_N = \mathrm{probv}(f,\mathrm{u} ,\mathcal{E}_{init}, N, \epsilon)$ \label{algline:probv1}
 \WHILE{$\rho^*_N \geq 0$}
    \STATE $\mathcal{E} = (\mathbf{1}_m + (i - 1) \, \Delta \mathbf{f}) \, \mathcal{E}_{init}$
    \STATE $\mathcal{E}_{temp} = (\mathbf{1}_m + i \, \Delta \mathbf{f}) \, \mathcal{E}_{init}$
    \STATE $\rho^*_N = \mathrm{probv}(f, \mathrm{u} ,\mathcal{E}_{temp}, N, \epsilon)$ \label{algline:probv2}
    \STATE $i \leftarrow i + 1$
 \ENDWHILE
 \IF{$i \neq 1$}
    \RETURN $\mathcal{E}$
 \ELSE
    \RETURN Reduce $\mathcal{E}_{init}$ as too large to verify
 \ENDIF
 \end{algorithmic}
 \end{algorithm}

In more detail, in line 2 of Alg.~\ref{alg:disturbanceset}, the initial expansion set is verified. If this initial set is not verifiable, a smaller initial set must be provided. Otherwise, the expansion set $\mathcal{E}_{temp}$, increased iteratively through $\Delta \mathbf{f}$ (line 5), is subsequently verified (line 6). Once the verification is not successful anymore, the algorithm terminates and returns the largest verified expansion set $\mathcal{E}$ (line 10). Note that we use an interval for $\mathcal{E}$ for simplicity. However, other set representation such as zonotopes would be possible.
To identify a more expressive expansion set, conformance checking~\cite{Roehm2019a} could be used.
 
\begin{figure}[tb]
	\vspace{0.4cm}
	\centering
	\resizebox{0.49\textwidth}{!}{%
			\begin{tikzpicture}[node distance=4cm, auto]  
			\tikzset{
				mynode/.style={rectangle,rounded corners,draw=black,draw=TUMpantone383,very thick, inner sep=1em, minimum size=3em, text centered, minimum width=8cm},
				myarrow/.style={
					->,
					thick,
					shorten <=2pt,
					shorten >=2pt,},
				mylabel/.style={text width=7em, text centered} 
			}  
			\node[mynode] (environment) [node distance=2.5cm, inner sep=0pt] {\LARGE{Environment}};
			\node(agent)  [below of=environment, minimum width=7.0cm, node distance=4.0cm, inner sep=0pt] {\LARGE{Agent}};
			\node (hp)[below = 1 cm of agent.south, node distance=0.0cm] {};
			\node[mynode, left of = hp, draw=TUMblue, minimum width=2.2cm, node distance=1.8cm, inner sep=7pt] (sc) {\makecell[c]{\Large{Safe Controller} \\ \Large{$u(x_t)$}}};
			\node[mynode, right of=hp, draw=TUMgray!40, minimum width=2.2cm, node distance=1.8cm, inner sep=7pt] (es) {\makecell[c]{\Large{Expansion Set} \\ \Large{$\mathcal{E}$}}};
			\begin{scope}[on background layer]
				\node [mynode,  fit={(agent) (es) (hp) (sc)}] (agentbox) {};
			\end{scope}
			\draw [myarrow, -latex] (environment.east) --++ (1,0) node(){} |- node[near start, left]{\makecell[l]{\Large{Observation} $o_t$ \\ \Large{Reward} $r_t$}} (agentbox);
			\draw [myarrow, -latex] (agentbox.west) --++ (-1,0) node(){} |-  node[near start, right]{\Large{Action} $a_{t}$} (environment.west);
			\end{tikzpicture} 
			\hspace{0.2cm}
			\begin{tikzpicture}[node distance=4cm, auto] 
			\draw[fill=white,draw=black](9,-2.0) rectangle (15,1.5);
			\draw[fill=white,draw=black](9,-6) rectangle (15,-2.5);
			\draw(9.4,1.1) node {\scalebox{1.5}{$\mathcal{X}$}};
			\draw(9.4,-2.9) node {\scalebox{1.5}{$\mathcal{U}$}};
			\draw [fill=black,draw=black](9.7,-0.8) circle (0.05) node (x0){};
			\draw(9.9,-1.1) node {\scalebox{1.4}{$x_0$}};
			
			\draw [fill=black,draw=black](10.3,-0.3) circle (0.05) node (x1){};
			\draw(10.4,-0.6) node {\scalebox{1.4}{$x_1$}};
			\path (x0) edge[bend left, ->, dashed, draw=TUMpantone383, thick] (x1);
			\draw [fill=black,draw=black](10,-0.1) circle (0.05) node (x1p){};
			\draw(9.8,0.3) node {\scalebox{1.4}{$x_1^\prime$}};
			\path (x0) edge[bend left, ->, dashed, draw=TUMblue, thick] (x1p);
			
			\draw [fill=black,draw=black](13,0.8) circle (0.05) node (x2){};
			\draw(13,0.4) node {\scalebox{1.4}{$x_2$}};
			\path (x1) edge[bend left, ->, dashed,draw=TUMpantone383, thick] (x2);
			\draw [fill=black,draw=black](12.5,-0.1) circle (0.05) node (x2p){};
			\draw(12.5,-0.4) node {\scalebox{1.4}{$x_2^\prime$}};
			\path (x1) edge[bend left, ->, dashed, draw=TUMblue, thick] (x2p);
			
			\draw [draw=TUMgray!40](11.3,-4.2) rectangle (12.7, -2.8);
			\draw (11.55,-3.9) node{\textcolor{TUMgray!40}{\scalebox{1.4}{$\mathcal{E}$}}};
			\draw [fill=TUMblue,draw=TUMblue](12,-3.5) circle (0.05);
			\draw (12.5,-3.2) node{\textcolor{TUMblue}{\scalebox{1.4}{$u(x_0)$}}};
			\draw [fill=TUMpantone383,draw=TUMpantone383](12.05,-3.8) circle (0.05);
			\draw (12.4,-3.9) node{\textcolor{TUMpantone383}{\scalebox{1.4}{$a_0$}}};
			
			\draw [draw=TUMgray!40](9.3,-5.5) rectangle (10.7,-4.1);
			\draw (9.6,-5.2) node{\textcolor{TUMgray!40}{\scalebox{1.4}{$\mathcal{E}$}}};
			\draw [fill=TUMblue,draw=TUMblue](10,-4.8) circle (0.05);
			\draw (10.5,-5.05) node{\textcolor{TUMblue}{\scalebox{1.4}{$u(x_1)$}}};
			\draw [fill=TUMpantone383,draw=TUMpantone383](10.05,-4.4) circle (0.05);
			\draw (10.4,-4.3) node{\textcolor{TUMpantone383}{\scalebox{1.4}{$a_1$}}};
		\end{tikzpicture} 
	} \vspace{-0.2 in}
    \caption{\ac{rl} within expansion set $\mathcal{E}$ around the safe controller $u(x)$ with trajectories for state space $\mathcal{X}$ and input space $\mathcal{U}$. The states $x$ are reached by the \ac{rl} actions, and $x^\prime$ are states that would be reached by the safe controller.}
    \label{fig:RL_process}
	\vspace{-0.2 in}
\end{figure}
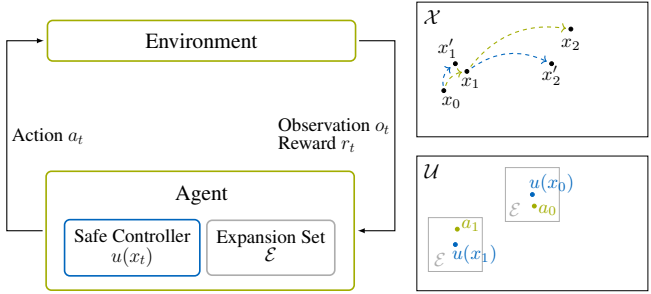

\spacing
\newidea{Step Two:} The second step in our \vSRL process is to learn a controller that improves for performance of the safe controller $\mathrm{u}$ we verified in the prior step. To constrain the learning based on the safe controller, we define a state-dependent action space $\mathcal{A}(x)$ around the safe control input $\mathrm{u}(x)$ inspired by continuous action masking \cite{Krasowski2022}:
\begin{equation}
\label{eq:action_set_with_disturbance_on_safe_control}
    \mathcal{A}(x) = \mathrm{u}(x) \oplus \mathcal{E},
\end{equation}
where $\oplus$ denotes the Minkowski sum. Based on Corollary~\ref{corr:prob_verify_cand_controller}, we can compute the probabilistic guarantee  (usually $\approx 1$) for the closed-loop system when perturbing the safe input $\mathrm{u}(x)$ with uniformly sampled noise within the expansion set $\mathcal{E}$. Therefore, if we continuously choose actions $a \in \mathcal{A}(x)$, our learned agent will with high probability yield safe behavior that also fulfills the probabilistic safety specification. Consequently, the agents can focus on optimizing for performance when learning within $\mathcal{E}$ around the safe controller. Thus, our three-step process delineates the safety (step one and three) and performance aspects (step two). This simplifies the reward and observation definitions as we only need to consider performance. 
Fig.~\ref{fig:RL_process} depicts this learning process and shows how the \ac{rl} agent can effect a system trajectory in the state space by changing the control input within $\mathcal{A}(x)$.

\spacing
\newidea{Step Three:} The third step is to verify the learned agent with a deterministic policy $\pi_a:\mathcal{X} \to \mathcal{U}$, to ensure the preservation of safety after optimizing for performance through learning. This is necessary since the learned agent (obtained by step two) will be different from the probabilistically verified safe controller with disturbance uniformly sampled from $\mathcal{E}$ (verified in step one). Thus, the probabilistic verification result is likely to hold but still needs to be verified for the learned agent. This verification, however, amounts to one more implementation of Theorem~\ref{thm:prob_verification}:  
\begin{corollary}
    Let the system dynamics be as per~\eqref{eq:basic_information} with a learned \ac{rl} agent $\pi_a:\mathcal{X} \to \mathcal{U}$ as controller, let the system safety specification $\psi$ have a robustness measure $\rho$ as per Definition~\ref{def:robustness}, and let $\mathcal{D}$ and $\rho^*_N$ be as in Corollary~\ref{corr:prob_verify_cand_controller}. Then for some $\epsilon \in [0,1]$, $\rho^*_N$ underperforms the $1-\epsilon$-th quartile of robustnesses achievable by the learned system with minimum confidence $1-(1-\epsilon)^N$.
    \begin{equation} \prob^N_{\uniform[\mathcal{X}]}\left[\prob_{\uniform[\mathcal{X}]}\left[\rho(\phi_{x_0}) \geq \rho^*_N\right] \geq 1-\epsilon \right] \geq 1-(1-\epsilon)^N.
    \end{equation}
\end{corollary}

\vspace{0.3cm}
\begin{proof}
    This is an application of Theorem~\ref{thm:prob_verification}.
\end{proof}

\section{Safe Evasion Task with Experimental Validation}\label{sec:casestudy}
To make our high-level approach more tractable, we define a problem with a safety specification that is more complex than a simple reach-avoid specification. Specifically, the mobile robot's task is to follow an optimal path to the goal while complying with a temporal logic safety specification -- whenever a collision is possible with the dynamic obstacle of the environment within the next few time steps, the mobile robot has to evade in a specified manner. Relevant examples for real-world applications where only a specific evasion is safe are: autonomous vehicles that have to overtake another traffic participant in a specific lane~\cite{Maierhofer2020.trafficrules} or autonomous vessels that have to perform specific collision avoidance maneuvers in order to be predictable for other ships~\cite{Torben2022.trafficrules}. We first describe the safety specification and \ac{rl} problem. Our experiments are conducted on the Robotarium~\cite{Pickem2017} and its simulation.  

\begin{figure*}
	\vspace{0.2cm}
	\centering
	\resizebox{0.8\textwidth}{!}{%
		\input{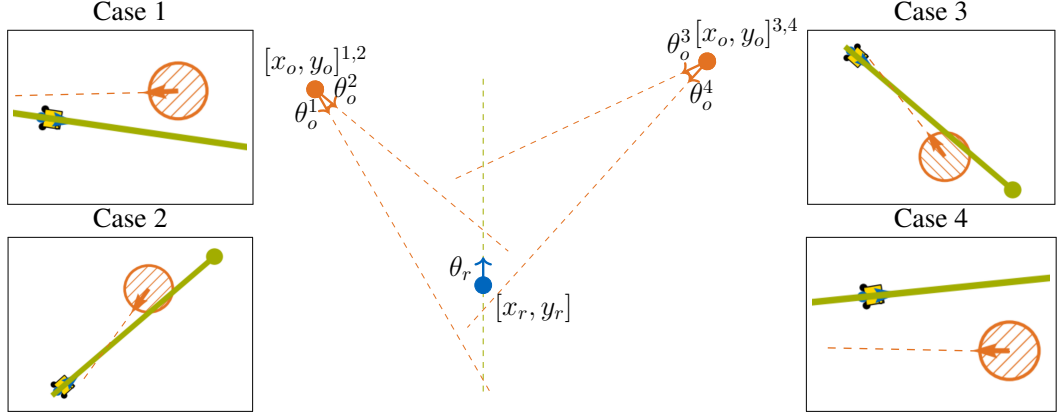}
	}
	\caption{Visualization of the four different relative orientations and position cases. The obstacle is depicted in orange and the robot in blue, and the case is indicated by the superscript. For cases 1 and 3, the robot should turn right (i.e., $sign = 1$) and for the cases 2 and 4, the robot should turn left (i.e., $sign = -1$).}
    \label{fig:turning_direction}
	\vspace{-0.2 in}
\end{figure*}

\subsection{Safety Specification}
The state of the robot is $r = [x_r, y_r, \theta_r, v_r] \in \mathbb{R}^4$ where $x_r$ and $y_r$ describe the position of the robot, $\theta_r$ is its orientation, and $v_r$ is its velocity aligned with its orientation. There is always one dynamic obstacle present and it is described by the state $o = [x_o, y_o, \theta_o, v_o] \in \mathbb{R}^4$. The time step is $\Delta t$. We assume a unicycle model for the robot for which control inputs $u$ are velocity and turning rate. Note that for the Robotarium, we provide the same control inputs and their robot controller generates the corresponding actuator signals.

Our safety specification is as follows: When the projected positions of the robot and any obstacle are closer than $\SI{0.4}{\meter}$ for any time steps within $\SI{1}{\second}$, then the robot should evade in a specific manner that depends on the relative positions and orientations of the obstacle and agent. To formalize the specification with \ac{stl}, we first need to define a few predicates and functions.

We can convert heading angles to unit vectors via $\a2v (\phi) = [\cos(\phi), \sin(\phi)]$. The rotation matrix for an angle $\alpha$ is 
\begin{equation}
    R_\alpha = \begin{bmatrix} \cos(-\alpha) & \sin(-\alpha) \\
-\sin(-\alpha) & \cos(-\alpha)
\end{bmatrix}.
\end{equation} 
The function $\sgn(x)$ returns $-1$ if $x < 0$, $1$ if $x > 0$ and $0$ otherwise. The minimum distance function $\mindistance(r,o,\Delta t)$ is defined as:
\begin{equation}
\begin{split}
    \mindistance(r,o, \Delta t) = \min_{t \in \{0,\Delta t, \dots, \SI{1}{\second}\}} &\Bigg( \norm[\Bigg]{ 
    \left(\begin{bmatrix}
        x_r \\
        y_r
    \end{bmatrix}
    + \a2v(\theta_r) v_r t \right) \\ 
    & - 
    \left(
    \begin{bmatrix}
        x_o \\
        y_o
    \end{bmatrix} + \a2v(\theta_o) v_o t \right)}_2 \Bigg),
    \end{split}
\end{equation}
where the time step size is $\Delta t=\SI{0.033}{\second}$. The minimum distance prediction assumes that the robot and the obstacle will move in their current directions with their current speeds, which is often a reasonable prediction. 
The predicate $\infront(r,o)$ checks if the obstacle is in the halfspace in front of the vehicle, \textit{i.e.} it evaluates to true iff $[x_o, y_o] \cdot \a2v(\theta_r) -  b_r \geq 0$ where $b_r$ is the offset.
Safe evasion is specified by the predicate:
\begin{equation}
    \evade(\dot{\theta}_r,\Delta \theta, sign) = \begin{cases}
        \text{True,} & \text{iff} \, (|\dot{\theta}_r| \leq \SI{1.5}{\radian \per \second} \\
        &\land \sgn(\dot{\theta}_r) = sign) \lor \\
        & ((\Delta \theta \geq \SI{0}{\radian} \\
        &\lor |\Delta \theta| \leq \SI{0.01}{\radian}) \\
        &\land \SI{0.01}{\radian \per \second} \geq |\dot{\theta}_r| ) \\
        \text{False,} &\text{otherwise,} \\
    \end{cases} 
\end{equation}
where $\Delta \theta \in [-\pi, \pi]$ is the orientation difference to the orientation perpendicular to the direction of the straight path between the initial state and goal in the direction of turning; $sign$ is $-1$ if the robot should turn to the left and $1$ in case the robot should turn to the right.  See Fig.~\ref{fig:turning_direction} for a depiction of the case-by-case evasion situations. Note that for our task specification with one non-reactive dynamic obstacle, the four cases are exhaustive for identifying the direction of evasion. To give an intuition, a safe evasion maneuver is that the robot turns until its orientation is perpendicular to the straight path between the initial state and goal. If this orientation is reached, the robot is no longer required to turn\footnote{No turning would be an impossible requirement due to the stochasticity needed for the input space, which includes the turning rate.} and continues driving away from the direct path between the initial state and goal to evade further. 

With these predicates and functions, the \ac{stl} formula is:
\begin{equation}
\begin{split}
\label{eq:STL_formula}
	G\Bigl((\infront(r,o) \land \mindistance(r,o,\Delta t) &\leq \SI{0.4}{\meter}) \implies \\ & \evade(\dot{\theta}_r,\Delta \theta, sign)\Bigr).
 \end{split}
\end{equation}
The robustness measure for this \ac{stl} formula is:
\begin{equation}
    \rho = \begin{cases}
        -1, & \text{iff eq. }\eqref{eq:STL_formula} = False\\
        \sum_{k=1}^{K} \perform(r_K, r_0, goal, K), & \text{otherwise},\\
    \end{cases}
\end{equation}
where $K$ is the length of the trajectory, $goal$ is the position of the goal, $r_K$ is the position of the robot at the end of the trajectory, $r_0$ is the initial position of the robot, and the performance objective is defined by
\begin{align*}
    \perform(r_K, r_0, goal,K)= & \max\left(\left(1 - \frac{\|r_K - goal\|_2}{\|r_0 - goal\|_2}\right), 0 \right)  \\  & + \left(1 - \frac{K}{K_{max}}\right),
\end{align*} 
with the maximal length of a trajectory $K_{max}=300$.
Note that although the objective of this safety specification is collision avoidance, which can also be achieved with other formal methods such as control barrier functions, these methods cannot easily be used to guarantee a specific avoidance behaviour as defined in the predicate $\evade$. Additionally, note that this particular safety specification can also be expressed with \ac{ltl}. However, as \ac{stl} is more expressive than \ac{ltl}, \ac{stl} can be utilized to express more complex specifications, \textit{e.g.} for marine traffic rules \cite{Torben2022.trafficrules}.

\subsection{Reinforcement Learning Problem}
We obtain the action space as described in~\eqref{eq:action_set_with_disturbance_on_safe_control}.
The reward function $\mathrm{R}$ is the difference between the achieved state with the \ac{rl} control input and only with the safe control input
\begin{equation}
    \mathrm{R} = r_{\mathrm{diff}} \cdot (\|goal - [x_{sc}, y_{sc}] \|_2 - \|goal - [x_{r}, y_{r}] \|_2),
\end{equation}
where $r_{\mathrm{diff}} \in \mathbb{R}_{+}$ scales the reward, and $[x_{sc}, y_{sc}]$ is the agent's position if the input from the safe controller would be used in the previous state, \textit{i.e.}  $a_{t-1} = u_{t-1}$.
Our agent can observe its relative position to the goal, the relative position to the closest point on the optimal path, the orientation difference to the optimal orientation calculated from the initial position to the goal, and the relative position to the obstacle. Note that this observation differs from the state.

We used Proximal Policy Optimization (PPO)~\cite{Schulman2017} as our \ac{rl} algorithm and scaled the reward $\mathrm{R}$ with the factor $r_{\mathrm{diff}}$ such that the reward is approximately between $-5$ and $5$ per episode. We identified the best hyperparameters for PPO with a small search of eight different configurations and three random seeds. However, all tested hyperparameters showed a converging behavior.
The hyperparameters that differ from the default PPO configuration of stable-baselines3~\cite{stable-baselines3} are the network architecture (two-layer multi-layer perceptron with 128 neurons in each layer), the value function coefficient (0.05), and the entropy coefficient (0.01). We train the agent for one million training steps.
\begin{figure}
    \vspace{0.3cm}
    \centering
  \subfloat[\label{fig:reward}]{%
       \includegraphics[width=0.4\textwidth]{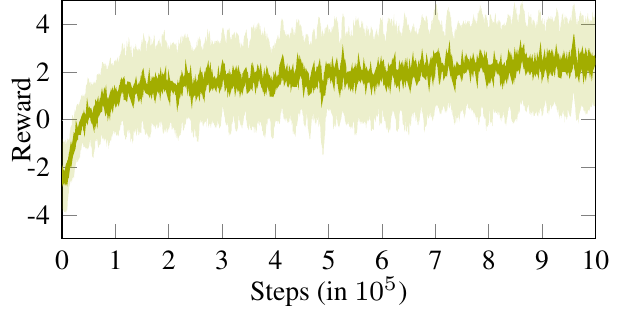}}\\
  \subfloat[\label{fig:action_usage}]{%
        \includegraphics[width=0.4\textwidth]{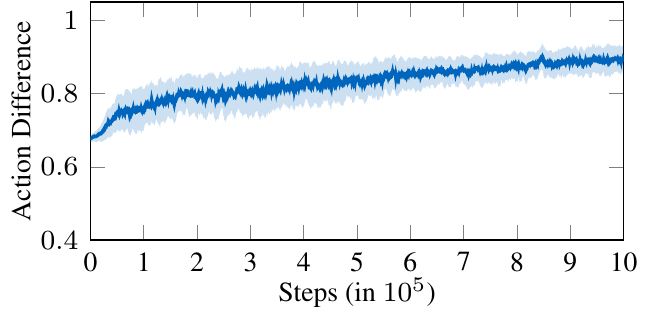}}
  \caption{Training curves for (a) reward function and (b) Euclidean norm difference between the learned agent's action and the safe control input scaled between 0 and 1. The action difference of a step is one if the agent selects an action on the border of the expansion set and zero if the agent does not alter the safe control input. The curves depict the mean and standard deviation for five different random seeds with the same hyperparameters.}
  \label{fig:training_results} 
  \vspace{-0.1cm}
\end{figure} 

\begin{figure*}
\vspace{0.3cm}
\centering{
\resizebox{0.9\textwidth}{!}{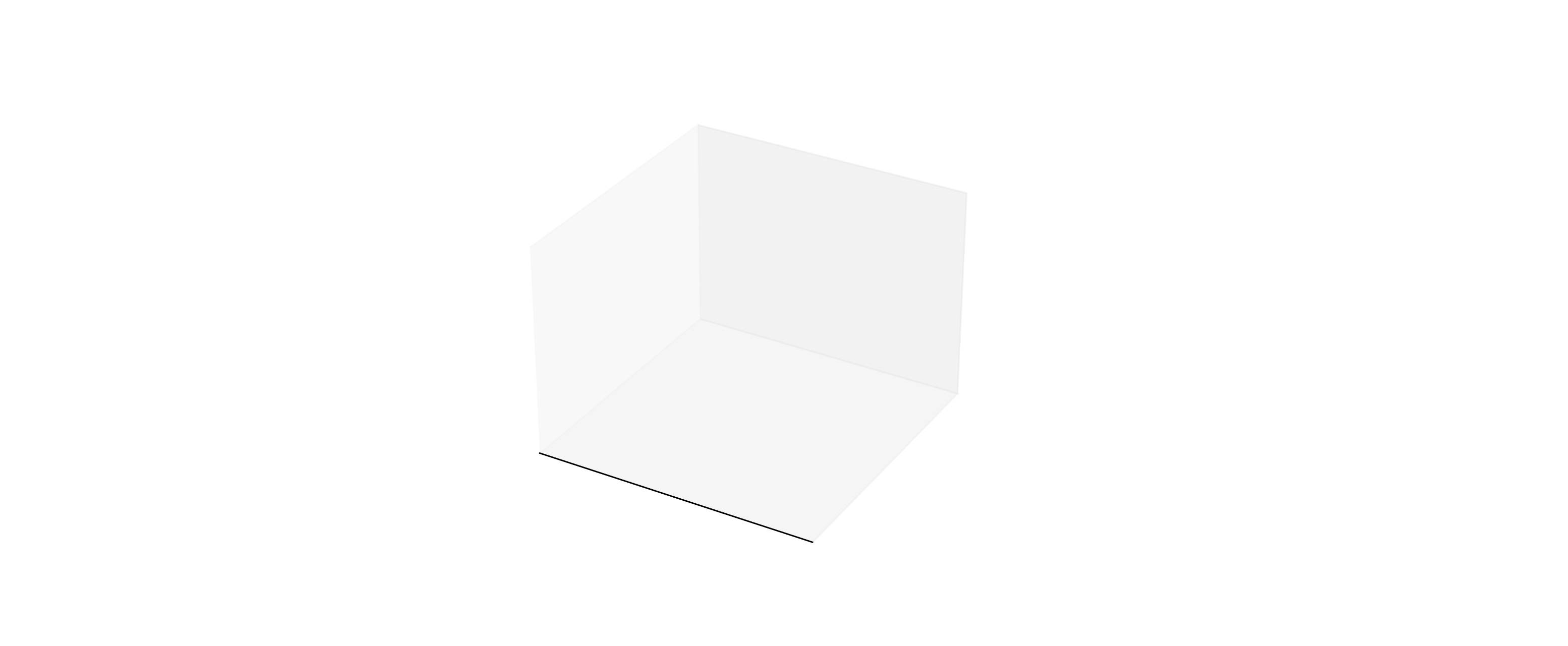}
\caption{Center: Robustness measure $\rho(s)$ histograms for 200 samples; Robotarium trajectories for different robustness values for learned agent (left top 0.114, bottom 1.62) and for safe controller (right top: 0.089, bottom: 1.21).}
\label{fig:results}
}
\end{figure*}

\spacing
\newidea{Verification of Safe Controller (Step 1):}
We want to verify whether our safe controller with uniformly sampled perturbations from the expansion set $\mathcal{E}$ achieves safe behavior with $95\%$ probability ($\epsilon = 0.05$),  \textit{i.e.} has a probabilistic cutoff $\rho^*_N > 0$ as per Theorem~\ref{thm:prob_verification} with $N=50$ samples. We apply Alg.~\ref{alg:disturbanceset} with $\mathcal{E}_{init} = [-0.0002, 0.0002] \si{\meter\per\second} \times [-0.005, 0.005] \si{\radian\per\second}$ and $\Delta \mathbf{f} = [10 , 1.0]$  and obtain $ \mathcal{E} = [-0.002, 0.002] \si{\meter\per\second} \times [-0.01, 0.01] \si{\radian\per\second}$.  Verifying the safe controller on the unicycle model with $N=50$ samples, yielded a probabilistic cutoff $\rho^*_{50} = 0.276$ (see Fig.~\ref{fig:results}).  This indicates that our chosen safe controller successfully exhibits safe behavior with $95\%$ probability, and we are $92.3\%$ confident in this statement --- probabilities were calculated via substitution in~\eqref{eq:prob_statement}.

\newidea{Reinforcement Learning (Step 2):}
We expect that if we define our action space as in~\eqref{eq:action_set_with_disturbance_on_safe_control}, the previous verification result will most likely hold for the learned agent. Fig.~\ref{fig:training_results} depicts the episode reward and action difference to the safe control input over the training steps. A reward above zero indicates that the agent performs better with respect to goal reaching than the safe controller. This is the case after approximately 50000 training steps. Additionally, the difference between the safe control input and the agent's action increases over the training towards the agent mainly selecting actions close to its action space boundary.

\newidea{Verification of Learned Agent (Step 3):}
Following the same verification procedure as for the stochastic safe controller yielded probabilistic cutoffs $\rho^*_{50} = 0.276$ for the safe controller without perturbations and $\rho^*_{50} = 0.221$ for the learned agent. Additionally, Fig.~\ref{fig:results} shows the robustness measure histograms for 200 samples. For the learned agent, the distribution is shifted to slightly higher robustness values (robustness measure mean is 0.78 and standard deviation is 0.32) in comparison to the deterministic safe controller (robustness measure mean is 0.76 and standard deviation is 0.35). Since positive robustness measure values encode faster goal reaching, our numerical results imply that the learned agent improves the performance of the deterministic safe controller, while still exhibiting the same probabilistic level of \ac{stl} specification compliance. The histogram for the stochastic safe controller has the same range and a similar shape as the histogram for the learned agent, which indicates that the verification result of the perturbed safe controller is, in fact, a good approximator of learned system behavior.

\newidea{Testing on Robotarium Robot:}
Although our implementation is in Python and did not focus on efficiency, it was not necessary to improve the computational efficiency to be real-time capable for the Robotarium robot \cite{Pickem2017}. This is because we mainly run the forward evaluation of the policy network and safe controller online, which are computationally lightweight, and there is no need to predict and incorporate the safety of actions online which is often computation heavy.  Fig.~\ref{fig:results} shows example trajectories from the Robotarium experiments\footnote{Video of example trajectories: \url{https://youtu.be/8WWmKfh6WSM}}. We observe that for high robustness values the robot only has to evade shortly or not at all. The main reason for low robustness values is that the robot has to evade and the final time for our experiment is reached before the robot can return to the optimal path and reach the goal.

\section{Discussion}\label{sec:discussion}
Our implementation is a proof of concept for our probabilistically \vSRL approach and shows that for the safe evasion task probabilistic safety guarantees are not jeopardized by improving the performance with \ac{rl}. Since other approaches using \ac{rl} with temporal logic for safety specifications tackle a different problem (see Sec.~\ref{sec:introduction}), we compare our approach only with the baseline of the safe controller's performance. A more complex problem would be autonomous driving on highways, where it is very hard to fulfill performance and safety specifications simultaneously during controller synthesis but often a safe controller exists. In the future such more complex problems should be investigated but are out of scope for this paper.

An important assumption of our approach is that a safe (black-box) controller is available. This assumption can be satisfied through the utilization of existing methods to synthesize safe controllers from temporal logic specifications~\cite{belta2019formal,raman2014model} or from expert data via imitation learning~\cite{Hussein2017.imitationlearning}. Note that, we only require this controller to satisfy a safety specification, \textit{e.g.} avoid obstacles, and operate within safe bounds. Furthermore, as in the case of our experiment, it can be relatively easy to implement a safe controller. In particular, our safe controller consists of a high-level algorithm that provides waypoints to prevent a collision or track the optimal path as the situation requires. A low-level controller then tracks these waypoints and provides control inputs for the unicycle model. Note that we do not need to know the architecture of the safe controller as our approach regards it as a black-box. 

\section{Conclusion}\label{sec:conclusion} 
The proposed three-step \vSRL approach achieves effective behavior that satisfies a desired probabilistic \ac{stl} specification. 
Our results on a safe evasion task show that the separation of safety and performance leads to lean and efficient learning and the probabilistic \ac{stl} guarantees can easily be verified. The architecture of our approach removes the necessity of online predictions for the safety of actions, which, as a consequence, reduces the computation time and allows us to effortlessly run the controller in real-time.

\bibliographystyle{IEEEtran}
\bibliography{bib_works}

\end{document}